\documentclass{article}
\usepackage{iclr2020_conference,times}

\usepackage{amsmath,amsfonts,bm}

\def\eqref#1{equation~\ref{#1}}

\def\1{\bm{1}}

\DeclareMathAlphabet{\mathsfit}{\encodingdefault}{\sfdefault}{m}{sl}
\SetMathAlphabet{\mathsfit}{bold}{\encodingdefault}{\sfdefault}{bx}{n}

\usepackage{graphicx}
\usepackage{subcaption}
\usepackage{xcolor}
\usepackage{hyperref}
\usepackage{url}
\usepackage{booktabs} 
\usepackage{accents}
\usepackage{xspace}

\newcommand*{\yoruba}{Yor\`ub\'a\xspace}

\title{Distant Supervision and Noisy Label Learning for Low Resource Named Entity Recognition: A Study on Hausa and \yoruba}

\iclrfinalcopy

\newcommand*\samethanks[1][\value{footnote}]{\footnotemark[#1]}

\author{David I. Adelani\textsuperscript{1}\thanks{Equal contribution.}, Michael A. Hedderich\textsuperscript{1}\samethanks, Dawei Zhu\textsuperscript{1}\samethanks, Esther van den Berg\textsuperscript{2}, Dietrich Klakow\textsuperscript{1} \\
\textsuperscript{1}Spoken Language Systems (LSV), Saarland University,
Saarland Informatics Campus, Germany\\
\textsuperscript{2}Institute for Computational Linguistics, Heidelberg University, Germany\\
\texttt{\{didelani,mhedderich,dzhu,dietrich.klakow\}@lsv.uni-saarland.de}
}

\begin{document}

\maketitle

\begin{abstract}
The lack of labeled training data has limited the development of natural language processing tools, such as named entity recognition, for many languages spoken in developing countries. Techniques such as distant and weak supervision can be used to create labeled data in a (semi-) automatic way. Additionally, to alleviate some of the negative effects of the errors in automatic annotation, noise-handling methods can be integrated. Pretrained word embeddings are another key component of most neural named entity classifiers. With the advent of more complex contextual word embeddings, an interesting trade-off between model size and performance arises. While these techniques have been shown to work well in high-resource settings, we want to study how they perform in low-resource scenarios. In this work, we perform named entity recognition for Hausa and \yoruba, two languages that are widely spoken in several developing countries. We evaluate different embedding approaches and show that distant supervision can be successfully leveraged in a realistic low-resource scenario where it can more than double a classifier's performance.
\end{abstract}

\section{Introduction}

Named Entity Recognition (NER) is a classification task that identifies words in a text that refer to entities (such as dates, person, organization and location names). It is a core task of natural language processing and a component for many downstream applications like search engines, knowledge graphs and personal assistants. For high-resource languages like English, this is a well-studied problem with complex state-of-the-art systems reaching close to or above 90\% F1-score on the standard datasets CoNLL03 \citep{ner/Baevski19SOTA} and Ontonotes \citep{ner/Akbik18SOTA}. In recent years, research has been extended to a larger pool of languages including those of developing countries \citep{LORELEI, ner/Zhang18ELISA, noise/Mayhew19, noise/Cao19}. Often, for these languages (like Hausa and \yoruba studied here), there exists a large population with access to digital devices and internet (and therefore digital text), but natural language processing (NLP) tools do not support them. 

One key reason is the absence of labeled training data required to train these systems. While manually labeled, gold-standard data is often only available in small quantities, it tends to be much easier to obtain large amounts of unlabeled text. Distant and weak supervision methods can then be used to create labeled data in a (semi-) automatic way. Using context \citep{distant/LimitsWeaklyInstagram, noise/Wang19}, external knowledge and resources \citep{data/Webvision, distant/Pan17WikiAnn}, expert rules \citep{distant/DataProgramming, distant/Ratner2019Snorkel} or self-training \citep{distant/ChenSelfTrain, noise/Paul19}, a corpus or dataset can be labeled quickly and cheaply. Additionally, a variety of noise-handling methods have been proposed to circumvent the negative effects that errors in this automatic annotation might have on the performance of a machine learning classifier.

In this work, we study two methods of distant supervision for NER: Automatic annotation rules and matching of lists of entities from an external knowledge source. While distant supervision has been successfully used for high resource languages, it is not straight forward that these also work in low-resource settings where the amount of available external information might be much lower. The knowledge graph of Wikidata e.g. contains 4 million person names in English while only 32 thousand such names are available in \yoruba, many of which are Western names.

Orthogonally to distant supervision, the pre-training of word embeddings is a key component for training many neural NLP models. A vector representation for words is built in an unsupervised fashion, i.e. on unlabeled text. Standard embedding techniques include Word2Vec \citep{embeddings/Word2Vec}, GloVe \citep{embeddings/GloVe} and FastText \citep{embeddings/FastText}. In the last two years, contextualized word embeddings have been proposed \citep{embeddings/Elmo, embeddings/GPT2, embeddings/BERT}. At the cost of having a much larger model size, these vector representations take the context of words into account and have been shown to outperform other embeddings in many tasks. In this study, we evaluate both types of representations.

The key questions we are interested in this paper are: How do NER models perform for Hausa and \yoruba, two languages from developing countries? Are distant-supervision techniques relying on external information also useful in low-resource settings? How do simple and contextual word embeddings trade-off in model size and performance?

\section{Background \& Methods}

\subsection{Languages}

\textbf{Hausa language} is the second most spoken indigenous language in Africa with over 40 million native speakers \citep{Ethnologue2019}, and one of the three major languages in Nigeria, along with Igbo and \yoruba. The language is native to the Northern part of Nigeria and the southern part of Niger, and it is widely spoken in West and Central Africa as a trade language in eight other countries: Benin, Ghana, Cameroon, Togo, Côte d'Ivoire, Chad, Burkina Faso, and Sudan. Hausa has several dialects but the one regarded as standard Hausa is the \textit{Kananci} spoken in the ancient city of Kano in Nigeria. Kananci is the dialect popularly used in many local (e.g VON news\footnote{\url{https://www.von.gov.ng/hausa/}}) and international news media 
such as BBC, VOA, DW and Radio France Internationale. Hausa is a tone language but the tones are often ignored in writings, the language is written in a modified Latin alphabet. Despite the popularity of Hausa as an important regional language in Africa and it's popularity in news media, it has very little or no labelled data for common NLP tasks such as text classification, named entity recognition and question answering.

\textbf{\yoruba language} is the third most spoken indigenous language in Africa after Swahilli and Hausa with over 35 million native speakers \citep{Ethnologue2019}. The language is native to the South-western part of Nigeria and the Southern part of Benin, and it is also spoken in other countries like Republic of Togo, Ghana, Côte d'Ivoire, Sierra Leone, Cuba and Brazil. 
\yoruba has several dialects but the written language has been standardized by the 1974 Joint Consultative Committee on Education~\citep{Asahiah2017}, it has 25 letters without the Latin characters  (c, q, v, x and z) and with additional characters ({\d e}, gb, {\d s} , {\d o}).
\yoruba is a tone language and the tones are represented as diacritics in written text, there are three tones in \yoruba namely low ( \textbackslash
),
mid (``$-$'') and 
high ($/$). The mid tone is usually ignored in writings. Often time articles written online including news articles\footnote{\url{https://www.von.gov.ng/yoruba/}, and \url{https://www.bbc.com/yoruba}} like BBC and VON ignore diacritics. Ignoring diacritics makes it difficult to identify or pronounce words except they are in a context. For example, \textit{ow\'{o}} (money), \textit{\d o}w{\d {\`{o}}} (broom), \textit{\`{o}}w{\`{o}} (business), \textit{\d {\`{o}}}w{\d {\`{o}}} (honour), \textit{\d o}w{\d {\'{o}} (hand), and \textit{\d {\`{o}}}w{\d{\'{o}}}} (group) will be mapped to \textit{owo} without diacritics. Similar to the Hausa language, there are few or no labelled datasets for NLP tasks.

\subsection{Datasets \& Embeddings}
The Hausa data used in this paper is part of the LORELEI\footnote{\url{https://www.darpa.mil/program/low-resource-languages-for-emergent-incidents}} language pack. It consists of Broad Operational Language Translation (BOLT) data gathered from news sites, forums, weblogs, Wikipedia articles and twitter messages. We use a split of 10k training and 1k test instances. Due to the Hausa data not being publicly available at the time of writing, we could only perform a limited set of experiments on it.

The \yoruba NER data used in this work is the annotated corpus of Global Voices news articles\footnote{\url{https://yo.globalvoices.org/}} recently released by \citet{alabi2019massive}. The dataset consists of 1,101 sentences (26k tokens) divided into 709 training sentences, 113 validation sentences and 279 test sentences based on 65\%/10\%/25\% split ratio. The named entities in the dataset are personal names (PER), organization (ORG), location (LOC) and date \& time (DATE). All other tokens are assigned a tag of "O".

For the \yoruba NER training, we make use of \yoruba FastText embeddings~\citep{alabi2019massive}  and multilingual-BERT\footnote{\url{https://github.com/google-research/bert/blob/master/multilingual.md}} that was trained on 104 languages including \yoruba. Instead of the original FastText embeddings \citet{embeddings/FastText}, we chose FastText embeddings trained on a multi-domain and high-quality dataset \citep{alabi2019massive} because it gave better word similarity scores.

\subsection{Distant and Weak Supervision}

In this work, we rely on two sources of distant supervision chosen for its ease of application:

\textbf{Rules} allow to apply the knowledge of domain experts without the manual effort of labeling each instance. They are especially suited for entities that follow specific patterns, like time phrases in text (see also \cite{distant/Stroetgen18Time}). We use them for the DATE entity. 
In Yoruba, date expressions are written with the keywords of 
``\textit{\d oj\d{\'{o}}}'' (day), ``\textit{o\d{s}\`{u}}'' (month), 
and ``\textit{\d od{\d{\'{u}}}}n'' (year). 
Similarly, time expressions are written with keywords such as ``\textit{w\'{a}k\`{a}tí}'' (hour), ``\textit{\`{i}\d{s}\d{\'{e}}j\'{u}} (minute) and ``\textit{\`{i}\d{s}\d{\'{e}}j\'{u}-aaya} (seconds). 
Relative date and time expressions are also written with keywords ``\textit{\d lod{\d{\'{u}}}}n'' (in the year), ``\textit{lo\d{s}\`{u}}'' (in the month), 
``\textit{l\d{o}s{\d{\`{e}}}}'' (in the week), 
``\textit{l\d oj\d{\'{o}}}'' (in the day). An example of a date expression is: 

\textit{``8th of December, 2018''}  in \yoruba translates to  ``\textit{\d oj\d{\'{o}} 8 o\d{s}\`{u} \d{O}p{\d{\`{e}}}, \d{o}d\'{u}n 2018}''

\textbf{Lists of Entities} can be obtained from a variety of sources like gazetteers, dictionaries, phone books, census data and Wikipedia categories \citep{distant/Ratinov2009}. In recent years, knowledge bases like Freebase and Wikidata have become another option to retrieve entity lists in a structured way. An entity list is created by extracting all names of that type from a knowledge source (e.g. all person names from Wikidata). If a word or token from the unlabeled text matches an entry in an entity list, it is assigned the corresponding label. Experts can add heuristics to this automatic labeling that improve the matching \citep{distant/Dembowski2017}. These include e.g. normalizing the grammatical form of words or filtering common false positives.

Another popular method for low-resource NER is the use of cross-lingual information \citep{ner/Rahimi2019Massively}. Alternatives to distant supervision are crowd-sourcing \citep{noise/Rehbein17} and non-expert annotations \citep{distant/Mayhew18Talen}.

\subsection{Learning With Noisy Labels}

The labels obtained through distant and weak supervision methods tend to contain a high amount of errors. In the Food101N dataset \citep{data/FoodN} around 20\% of the automatically obtained labels are incorrect while for Clothing1M \citep{data/Xiao2015Clothing1M} the noise rate is more than 60\%. Learning with this additional, noisily labeled data can result in lower classification performance compared to just training on a small set of clean labels (cf. e.g. \citet{noise/Fang16POS}). A variety of techniques have been proposed to handle label noise like modelling the underlying noise process \citep{noise/Lange19} and filtering noisy instances \citep{Yang18Distantly, noise/FilteringNguyen}. \citet{noise/survey/Frenay2014} gives an in-depth introduction into this field and \citet{noise/survey/Algan19} survey more recent approaches, focusing on the vision domain.

In this work, we experiment with three noise handling techniques. The approach by \citet{noise/Bekker16} estimates a noise channel using the EM algorithm. It treats all labels as possibly noisy and does not distinguish between a clean and a noisy part of the data. In contrast, the method by \citet{noise/Hedderich18} leverages the existence of a small set of gold standard labels, something that - in our experience - is often available even in low resource settings. Having such a small set of clean labels is beneficial both for the main model itself as well as for the noise handling technique. Both approaches model the relationship between clean and noisy labels using a confusion matrix. This allows adapting the noisy to the clean label distribution during training. For a setting with 5 labels, it only requires $5^2=25$ additional parameters to estimate which could be beneficial when only few training data is available. The technique by \citet{noise/Veit17} (adapted to NER by \citet{noise/Hedderich18}) learns a more complex neural network to clean the noisy labels before training with them. It also takes the features into account when cleaning the noise and it might, therefore, be able to model more complex noise processes. All three techniques can be easily added to the existing standard neural network architectures for NER. 

\section{Models \& Experimental Settings}

\textbf{Hausa} Distant supervision on Hausa was performed using lists of person names extracted from Wikipedia data. 
Since we had limited access to the data, we tested a simplified binary NER-tagging setting (PERSON-tags only). As a base model, we used a Bi-LSTM model developed for Part-of-Speech tagging \citep{noise/plank}. For noise handling, we apply the \textbf{Noise Channel} model by \citet{noise/Bekker16}. 

\textbf{\yoruba} For \yoruba, the entity lists were created by extracting person, location and organization entities from Wikidata in English and \yoruba.  Additionally, a list of person names in Nigeria was obtained from a \yoruba Name website\footnote{\url{http://www.yorubaname.com/}} (8,365 names) and list of popular Hausa, Igbo, Fulani and \yoruba people on Wikipedia (in total 9,241 names). As manual heuristic, a minimum name length of 2 was set for extraction of PER (except for Nigerian names), LOC and ORG. The Nigerian names were set to include names with a minimum length of 3. 
For the DATE label, a native \yoruba speaker wrote some annotation rules using 11 ``date keywords'' (``\textit{\d oj\d{\'{o}}}'',  ``\textit{\d{o}s{\d{\`{e}}}}'', ``\textit{os{\d{\`{u}}}}'', ``\textit{\d od{\d{\'{u}}}}n'', ``\textit{w\'{a}k\`{a}tí}'' , ``\textit{\d lod{\d{\'{u}}}}n'', ``\textit{\d lod{\d{\'{u}}}}n-un'', ``\textit{\d od{\d{\'{u}}}}n-un'' ``\textit{l\d{o}s{\d{\`{e}}}}'' , ``\textit{l\d oj\d{\'{o}}}'', ``
\textit{aago}'') following these two criteria: (1) A token is a date keyword or follows a date keyword in a sequence. (2) A token is a digit. For \yoruba, we evaluate four settings with different amounts of clean data, namely 1k, 2k, 4k and the full dataset. As distantly supervised data with noisy labels, the full dataset is used. Additionally, 19,559 words from 18 articles of the Global News Corpus (different from the articles in the training corpus) were automatically annotated.

The \textbf{Bi-LSTM} model consists of a Bi-LSTM layer followed by a linear layer to extract input features. The Bi-LSTM layer has a 300-dimensional hidden state for each direction. For the final classification, an additional linear layer is added to output predicted class distributions. For noise handling, we experiment with the \textbf{Confusion Matrix} model by \cite{noise/Hedderich18} and the \textbf{Cleaning} model by \cite{noise/Veit17}. We repeat all the Bi-LSTM experiments 20 times and report the average F1-score (following the approach by \cite{data/CoNLL03}) and the standard error.

The \textbf{BERT} model is obtained by \textit{fine-tuning} the pre-trained BERT embeddings on NER data with an additional untrained CRF  classifier. We fine-tuned all the parameters of BERT including that of the CRF end-to-end. This has been shown to give better performance than using word features extracted from BERT to train a classifier~\citep{embeddings/BERT}. The evaluation result is obtained as an average of 5 runs, we report the F1-score and the standard error in the result section.

\section{Results}

\begin{figure}[t]
    
    \begin{subfigure}{0.49\textwidth}
        \centering
        \includegraphics[width=0.8\textwidth]{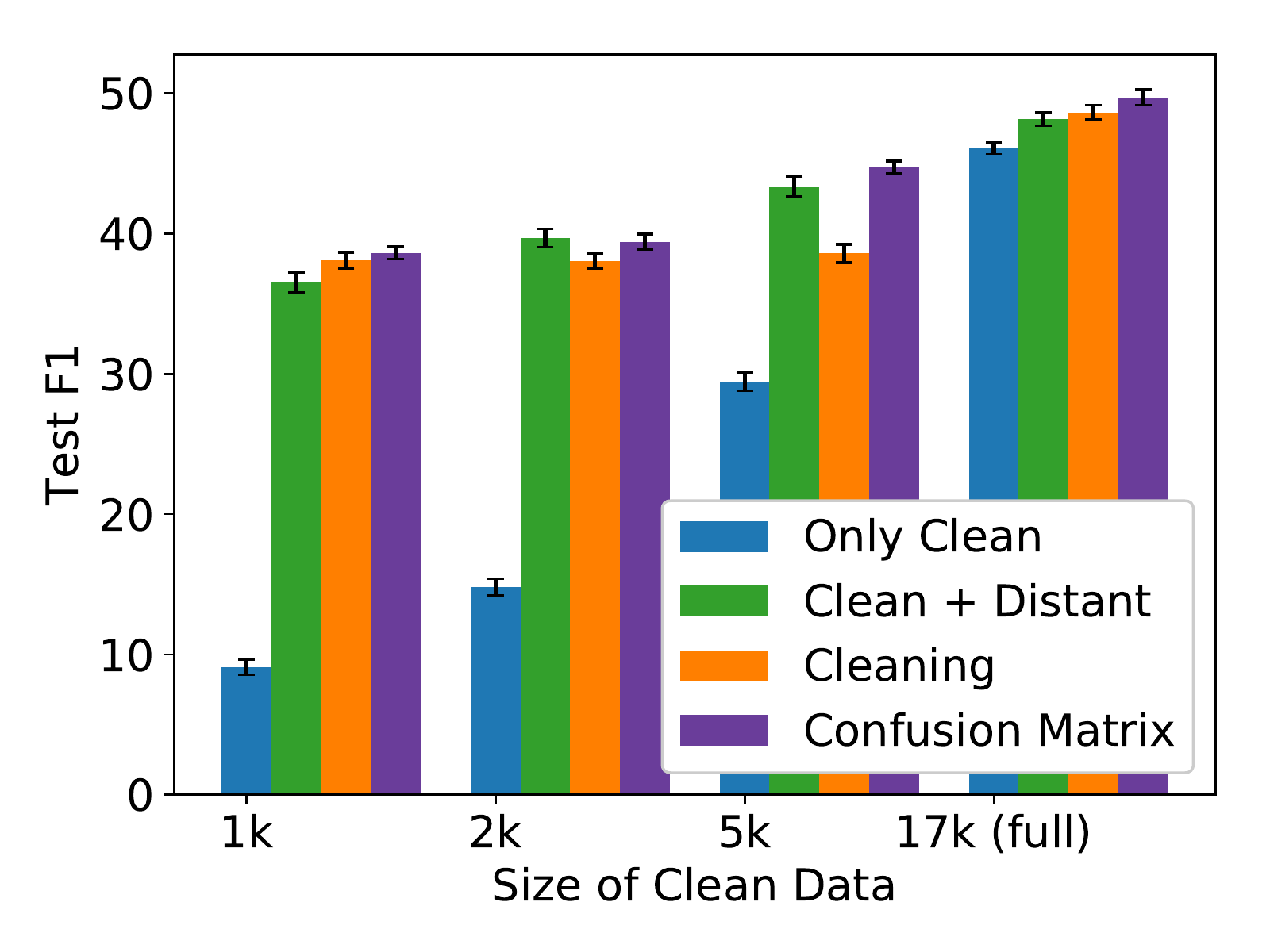}
        \caption{Bi-LSTM} \label{fig:yoruba_Bi-LSTM}
    \end{subfigure}
    \begin{subfigure}{0.49\textwidth}
        \centering
        \includegraphics[width=0.8\textwidth]{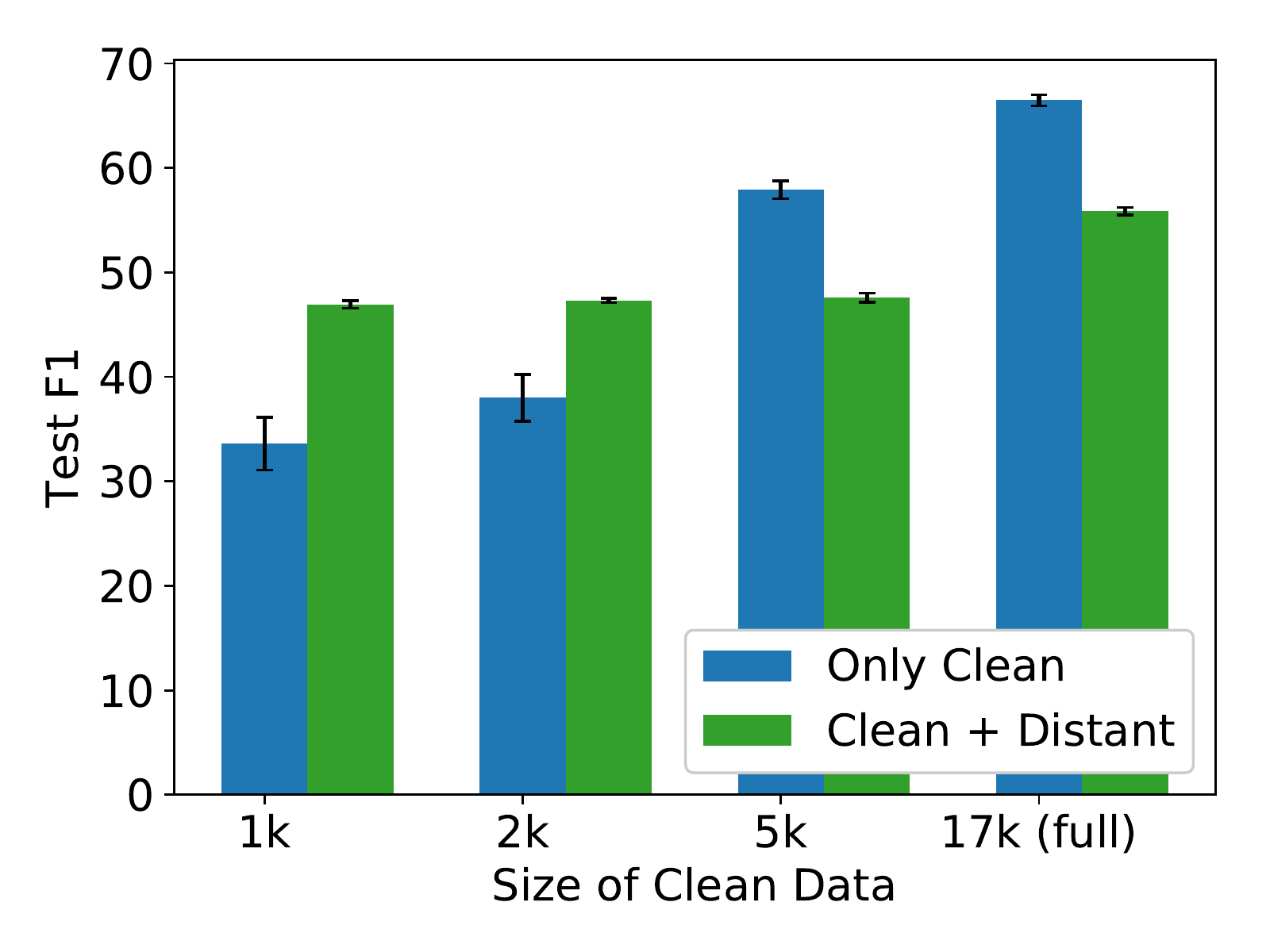}
        \caption{BERT} \label{fig:yoruba_bert}
    \end{subfigure}
    \caption{F1-scores and standard error for \yoruba. \label{fig:yoruba_results}}
\end{figure}

\begin{table}[t]
    \centering
    \scriptsize
    \begin{subtable}[t]{0.35\textwidth}
        \caption{Performance of the Bi-LSTM on Hausa (PER only) \label{tab:hausa_results}}
        \centering
        \begin{tabular}{p{1.5cm}|c|c|c}
            \toprule
            Data & Precision & Recall & F1\\
            \midrule 
            100\% Clean & 79 & 45 & 57 \\ 
            50\% Clean + 50\% Distant & 59 & 38 & 42 \\
            Noise Channel & 57 & 47 & 51  \\
            \bottomrule
        \end{tabular}
    \end{subtable}
    \begin{subtable}[t]{0.3\textwidth}
        \caption{Distant Supervision Quality (on test) \label{tab:distant_quality_all}}
        \centering
        \begin{tabular}{c|c|c|c}
            \toprule
            Method & P & R & F1\\
            \midrule
            Overall & 41 & 41 & 41 \\
            PER & 36 & 16 & 22\\
            LOC & 73 & 54 & 62\\
            ORG & 17 & 29 & 22\\
            DATE & 37 & 66 & 48 \\
            \bottomrule
        \end{tabular}
        \vspace{-2mm}
    \end{subtable}
    \begin{subtable}[t]{0.3\textwidth}
        \caption{Distant Supervision: Coverage of Wikidata for person names (on train) \label{tab:distant_quality_per}}
        \centering
        \begin{tabular}{c|c|c|c}
            \toprule
            Method & P & R & F1\\
            \midrule
            Wikidata & 50 & 25 & 33 \\
            + \yoruba  Names & 47 & 25 & 33 \\
            + Nigerian Names & 55 & 39 & 46 \\
            \bottomrule
        \end{tabular}
    \end{subtable}
\end{table}

The results for Hausa are given in Table \ref{tab:hausa_results}. Training with a mix of 50\% clean and 50\% distantly-supervised data performs 15 F1-score points below using the whole 100\% clean data which is to be expected due to the lower quality of the distantly-supervised labels. Using the Noise Channel closes half of this gap. Due to the limited availability of the dataset, we could unfortunately not investigate this further, but it shows already the benefits that are possible through noise-handling.

An evaluation of the distant supervision for \yoruba is given in Table \ref{tab:distant_quality_all}. The quality of the automatically annotated labels differs between the classes. Locations perform better than person and organization names, probably due to locations being less diverse and better covered in Wikidata. With simple date rules, we obtain already a 48\% F1-score. This shows the importance of leveraging the knowledge of native speakers in automatic annotations. Overall a decent annotation can be obtained by the distant supervision and it even outperforms some of the actual machine learning models in the low-resource setting. Table \ref{tab:distant_quality_per} compares using only Wikidata as data source versus adding additional, manually obtained lists of person names. While adding a list of \yoruba names only improves recall slightly, the integration of Nigerian names helps to boost recall by 13 points.

The experimental results for \yoruba are given in Figure \ref{fig:yoruba_results}. The setting differs from the experiments with Hausa in that there is a small clean training set and \textit{additional}, distantly-supervised data. For the Bi-LSTM model, adding distantly-supervised labels always helps. In the low-resource settings with 1k and 2k labeled data, it more than doubles the performance. Handling the noise in the distant supervision can result in slight improvements. The noise-cleaning approach struggles somewhat while the confusion matrix architecture does give better results in the majority of the scenarios. Training on 5k labeled data with distantly supervised data and noise handling, one can obtain a performance close to using the full 17k manually labeled token.

The Bi-LSTM model has 1.50 million parameters (1.53 million for the cleaning model), while BERT has 110 million parameters. There is a clear trade-off between model size and performance. The BERT model is 70 times larger and obtains consistently better results due to its more complex, contextual embeddings pretrained on more data. Still, the F1-score also drops nearly half for the BERT model in the 1k setting compared to the full dataset. For 1k and 2k labeled data, the distant supervision helps to improve the model's performance. However, once the model trained only on clean data reaches a higher F1-score than the distant supervision technique, the model trained on clean and distantly-supervised data deteriorates. This suggests that the BERT model overfits too much on the noise in the distant supervision.

\section{Conclusion}

In this study, we analysed distant supervision techniques and label-noise handling for NER in Hausa and \yoruba, two languages from developing countries. We showed that they can be successfully leveraged in a realistic low-resource scenario to double a classifier's performance. If model size is not a constraint, the more complex BERT model clearly outperforms the smaller Bi-LSTM architecture. Nevertheless, there is still a large gap between the best performing model on \yoruba with 66 F1-score and the state-of-the-art in English around 90.

We see several interesting follow-ups to these evaluations. In the future, we want to evaluate if noise handling methods can also allow the more complex BERT model to benefit from distant supervision. Regarding the model complexity, it would be interesting to experiment with more compact models like DistilBERT \citep{embeddings/DistilBERT} that reach a similar performance with a smaller model size for high-resource settings. In general, we want to investigate more in-depth the trade-offs between model complexity and trainability in low-resource scenarios.

\section*{Acknowledgments}
The experiments on Hausa were possible thanks to the collaboration with Florian Metze and CMU as part of the LORELEI project. Gefördert durch die Deutsche Forschungsgemeinschaft (DFG) – Projektnummer 232722074 – SFB 1102 / Funded by the Deutsche Forschungsgemeinschaft (DFG, German Research Foundation) – Project-ID 232722074 – SFB 1102, the EU-funded Horizon 2020 projects ROXANNE under grant agreement No. 833635 and COMPRISE (\texttt{http://www.compriseh2020.eu/}) under grant agreement No. 3081705.

\bibliography{iclr2020_conference}

\begin{thebibliography}{43}
\providecommand{\natexlab}[1]{#1}
\providecommand{\url}[1]{\texttt{#1}}
\expandafter\ifx\csname urlstyle\endcsname\relax
  \providecommand{\doi}[1]{doi: #1}\else
  \providecommand{\doi}{doi: \begingroup \urlstyle{rm}\Url}\fi

\bibitem[Akbik et~al.(2018)Akbik, Blythe, and Vollgraf]{ner/Akbik18SOTA}
Alan Akbik, Duncan Blythe, and Roland Vollgraf.
\newblock Contextual string embeddings for sequence labeling.
\newblock In \emph{Proceedings of the 27th International Conference on
  Computational Linguistics}, pp.\  1638--1649, Santa Fe, New Mexico, USA,
  August 2018. Association for Computational Linguistics.
\newblock URL \url{https://www.aclweb.org/anthology/C18-1139}.

\bibitem[Alabi et~al.(2020)Alabi, Amponsah-Kaakyire, Adelani, and
  Espa{\~n}a-Bonet]{alabi2019massive}
Jesujoba~O Alabi, Kwabena Amponsah-Kaakyire, David~I Adelani, and Cristina
  Espa{\~n}a-Bonet.
\newblock Massive vs. curated word embeddings for low-resourced languages. the
  case of {Yor$\backslash$ub$\backslash$'a} and {Twi}.
\newblock In \emph{LREC}, 2020.

\bibitem[Algan \& Ulusoy(2019)Algan and Ulusoy]{noise/survey/Algan19}
G{\"{o}}rkem Algan and Ilkay Ulusoy.
\newblock Image classification with deep learning in the presence of noisy
  labels: {A} survey.
\newblock \emph{CoRR}, abs/1912.05170, 2019.

\bibitem[Asahiah et~al.(2017)Asahiah, Odejobi, and Adagunodo]{Asahiah2017}
F.~O. Asahiah, O.~A. Odejobi, and E.~R. Adagunodo.
\newblock Restoring tone-marks in standard yoruba electronic text: Improved
  model.
\newblock \emph{Computer Science}, 18\penalty0 (3):\penalty0 301--315, 2017.

\bibitem[Baevski et~al.(2019)Baevski, Edunov, Liu, Zettlemoyer, and
  Auli]{ner/Baevski19SOTA}
Alexei Baevski, Sergey Edunov, Yinhan Liu, Luke Zettlemoyer, and Michael Auli.
\newblock Cloze-driven pretraining of self-attention networks.
\newblock In \emph{Proceedings of the 2019 Conference on Empirical Methods in
  Natural Language Processing and the 9th International Joint Conference on
  Natural Language Processing, {EMNLP-IJCNLP} 2019, Hong Kong, China, November
  3-7, 2019}, pp.\  5359--5368, 2019.
\newblock \doi{10.18653/v1/D19-1539}.
\newblock URL \url{https://doi.org/10.18653/v1/D19-1539}.

\bibitem[Bekker \& Goldberger(2016)Bekker and Goldberger]{noise/Bekker16}
Alan~Joseph Bekker and Jacob Goldberger.
\newblock Training deep neural-networks based on unreliable labels.
\newblock In \emph{2016 {IEEE} International Conference on Acoustics, Speech
  and Signal Processing, {ICASSP} 2016, Shanghai, China, March 20-25, 2016},
  pp.\  2682--2686, 2016.
\newblock \doi{10.1109/ICASSP.2016.7472164}.
\newblock URL \url{https://doi.org/10.1109/ICASSP.2016.7472164}.

\bibitem[Bojanowski et~al.(2016)Bojanowski, Grave, Joulin, and
  Mikolov]{embeddings/FastText}
Piotr Bojanowski, Edouard Grave, Armand Joulin, and Tomas Mikolov.
\newblock Enriching word vectors with subword information.
\newblock \emph{arXiv preprint arXiv:1607.04606}, 2016.

\bibitem[Cao et~al.(2019)Cao, Hu, Chua, Liu, and Ji]{noise/Cao19}
Yixin Cao, Zikun Hu, Tat{-}Seng Chua, Zhiyuan Liu, and Heng Ji.
\newblock Low-resource name tagging learned with weakly labeled data.
\newblock In \emph{Proceedings of the 2019 Conference on Empirical Methods in
  Natural Language Processing and the 9th International Joint Conference on
  Natural Language Processing, {EMNLP-IJCNLP} 2019, Hong Kong, China, November
  3-7, 2019}, pp.\  261--270, 2019.
\newblock \doi{10.18653/v1/D19-1025}.
\newblock URL \url{https://doi.org/10.18653/v1/D19-1025}.

\bibitem[Chen et~al.(2018)Chen, Zhang, and Gao]{distant/ChenSelfTrain}
Chenhua Chen, Yue Zhang, and Yuze Gao.
\newblock Learning how to self-learn: Enhancing self-training using neural
  reinforcement learning.
\newblock In \emph{2018 International Conference on Asian Language Processing,
  {IALP} 2018, Bandung, Indonesia, November 15-17, 2018}, pp.\  25--30, 2018.
\newblock \doi{10.1109/IALP.2018.8629107}.
\newblock URL \url{https://doi.org/10.1109/IALP.2018.8629107}.

\bibitem[Christianson et~al.(2018)Christianson, Duncan, and
  Onyshkevych]{LORELEI}
Caitlin Christianson, Jason Duncan, and Boyan~A. Onyshkevych.
\newblock Overview of the {DARPA} {LORELEI} program.
\newblock \emph{Machine Translation}, 32\penalty0 (1-2):\penalty0 3--9, 2018.
\newblock \doi{10.1007/s10590-017-9212-4}.
\newblock URL \url{https://doi.org/10.1007/s10590-017-9212-4}.

\bibitem[Dembowski et~al.(2017)Dembowski, Wiegand, and
  Klakow]{distant/Dembowski2017}
Julia Dembowski, Michael Wiegand, and Dietrich Klakow.
\newblock Language independent named entity recognition using distant
  supervision.
\newblock In \emph{Proceedings of Language and Technology Conference}, 2017.

\bibitem[Devlin et~al.(2019)Devlin, Chang, Lee, and Toutanova]{embeddings/BERT}
Jacob Devlin, Ming{-}Wei Chang, Kenton Lee, and Kristina Toutanova.
\newblock {BERT:} pre-training of deep bidirectional transformers for language
  understanding.
\newblock In \emph{Proceedings of the 2019 Conference of the North American
  Chapter of the Association for Computational Linguistics: Human Language
  Technologies, {NAACL-HLT} 2019, Minneapolis, MN, USA, June 2-7, 2019, Volume
  1 (Long and Short Papers)}, pp.\  4171--4186, 2019.
\newblock \doi{10.18653/v1/n19-1423}.
\newblock URL \url{https://doi.org/10.18653/v1/n19-1423}.

\bibitem[Eberhard et~al.(2019)Eberhard, Simons, and (eds.)]{Ethnologue2019}
David~M. Eberhard, Gary~F. Simons, and Charles D.~Fennig (eds.).
\newblock Ethnologue: Languages of the world. twenty-second edition., 2019.
\newblock URL \url{http://www.ethnologue.com}.

\bibitem[Fang \& Cohn(2016)Fang and Cohn]{noise/Fang16POS}
Meng Fang and Trevor Cohn.
\newblock Learning when to trust distant supervision: An application to
  low-resource pos tagging using cross-lingual projection.
\newblock In \emph{Proceedings of the 20th SIGNLL Conference on Computational
  Natural Language Learning}, 2016.
\newblock \doi{10.18653/v1/K16-1018}.

\bibitem[{Frenay} \& {Verleysen}(2014){Frenay} and
  {Verleysen}]{noise/survey/Frenay2014}
B.~{Frenay} and M.~{Verleysen}.
\newblock Classification in the presence of label noise: A survey.
\newblock \emph{IEEE Transactions on Neural Networks and Learning Systems},
  25\penalty0 (5):\penalty0 845--869, 2014.
\newblock \doi{10.1109/TNNLS.2013.2292894}.

\bibitem[Hedderich \& Klakow(2018)Hedderich and Klakow]{noise/Hedderich18}
Michael~A. Hedderich and Dietrich Klakow.
\newblock Training a neural network in a low-resource setting on automatically
  annotated noisy data.
\newblock In \emph{Proceedings of the Workshop on Deep Learning Approaches for
  Low-Resource {NLP}}. Association for Computational Linguistics, July 2018.

\bibitem[Lange et~al.(2019)Lange, Hedderich, and Klakow]{noise/Lange19}
Lukas Lange, Michael~A. Hedderich, and Dietrich Klakow.
\newblock Feature-dependent confusion matrices for low-resource ner labeling
  with noisy labels.
\newblock In \emph{Proceedings of the 2019 Conference on Empirical Methods in
  Natural Language Processing and the 9th International Joint Conference on
  Natural Language Processing (EMNLP-IJCNLP)}, pp.\  3545--3550. Association
  for Computational Linguistics, 2019.
\newblock \doi{10.18653/v1/D19-1362}.

\bibitem[Lee et~al.(2018)Lee, He, Zhang, and Yang]{data/FoodN}
Kuang{-}Huei Lee, Xiaodong He, Lei Zhang, and Linjun Yang.
\newblock Cleannet: Transfer learning for scalable image classifier training
  with label noise.
\newblock In \emph{2018 {IEEE} Conference on Computer Vision and Pattern
  Recognition, {CVPR} 2018, Salt Lake City, UT, USA, June 18-22, 2018}, pp.\
  5447--5456, 2018.
\newblock \doi{10.1109/CVPR.2018.00571}.

\bibitem[Li et~al.(2017)Li, Wang, Li, Agustsson, and Gool]{data/Webvision}
Wen Li, Limin Wang, Wei Li, Eirikur Agustsson, and Luc~Van Gool.
\newblock Webvision database: Visual learning and understanding from web data.
\newblock \emph{CoRR}, abs/1708.02862, 2017.

\bibitem[Mahajan et~al.(2018)Mahajan, Girshick, Ramanathan, He, Paluri, Li,
  Bharambe, and van~der Maaten]{distant/LimitsWeaklyInstagram}
Dhruv Mahajan, Ross Girshick, Vignesh Ramanathan, Kaiming He, Manohar Paluri,
  Yixuan Li, Ashwin Bharambe, and Laurens van~der Maaten.
\newblock Exploring the limits of weakly supervised pretraining.
\newblock In Vittorio Ferrari, Martial Hebert, Cristian Sminchisescu, and Yair
  Weiss (eds.), \emph{Computer Vision -- ECCV 2018}, pp.\  185--201. Springer
  International Publishing, 2018.

\bibitem[Mayhew \& Roth(2018)Mayhew and Roth]{distant/Mayhew18Talen}
Stephen Mayhew and Dan Roth.
\newblock Talen: Tool for annotation of low-resource entities.
\newblock In \emph{Proceedings of ACL 2018: System Demonstrations}, 2018.
\newblock URL \url{http://aclweb.org/anthology/P18-4014}.

\bibitem[Mayhew et~al.(2019)Mayhew, Chaturvedi, Tsai, and Roth]{noise/Mayhew19}
Stephen Mayhew, Snigdha Chaturvedi, Chen-Tse Tsai, and Dan Roth.
\newblock Named entity recognition with partially annotated training data.
\newblock In \emph{Proceedings of the 23rd Conference on Computational Natural
  Language Learning (CoNLL)}, pp.\  645--655, Hong Kong, China, November 2019.
  Association for Computational Linguistics.
\newblock \doi{10.18653/v1/K19-1060}.
\newblock URL \url{https://www.aclweb.org/anthology/K19-1060}.

\bibitem[Mikolov et~al.(2013)Mikolov, Chen, Corrado, and
  Dean]{embeddings/Word2Vec}
Tomas Mikolov, Kai Chen, Greg Corrado, and Jeffrey Dean.
\newblock Efficient estimation of word representations in vector space.
\newblock In \emph{1st International Conference on Learning Representations,
  {ICLR} 2013, Scottsdale, Arizona, USA, May 2-4, 2013, Workshop Track
  Proceedings}, 2013.
\newblock URL \url{http://arxiv.org/abs/1301.3781}.

\bibitem[Nguyen et~al.(2020)Nguyen, Mummadi, Ngo, Nguyen, Beggel, and
  Brox]{noise/FilteringNguyen}
T.~Nguyen, C.~K. Mummadi, T.~P.~N. Ngo, T.~H.~P. Nguyen, L.~Beggel, and
  T.~Brox.
\newblock Self: learning to filter noisy labels with self-ensembling.
\newblock In \emph{International Conference on Learning Representations
  (ICLR)}, 2020.

\bibitem[Pan et~al.(2017)Pan, Zhang, May, Nothman, Knight, and
  Ji]{distant/Pan17WikiAnn}
Xiaoman Pan, Boliang Zhang, Jonathan May, Joel Nothman, Kevin Knight, and Heng
  Ji.
\newblock Cross-lingual name tagging and linking for 282 languages.
\newblock In \emph{Proceedings of the 55th Annual Meeting of the Association
  for Computational Linguistics (Volume 1: Long Papers)}, pp.\  1946--1958,
  Vancouver, Canada, July 2017. Association for Computational Linguistics.
\newblock \doi{10.18653/v1/P17-1178}.
\newblock URL \url{https://www.aclweb.org/anthology/P17-1178}.

\bibitem[Paul et~al.(2019)Paul, Singh, Hedderich, and Klakow]{noise/Paul19}
Debjit Paul, Mittul Singh, Michael~A. Hedderich, and Dietrich Klakow.
\newblock Handling noisy labels for robustly learning from self-training data
  for low-resource sequence labeling.
\newblock In \emph{Proceedings of the 2019 Conference of the North {A}merican
  Chapter of the Association for Computational Linguistics: Student Research
  Workshop}, pp.\  29--34, Minneapolis, Minnesota, June 2019. Association for
  Computational Linguistics.
\newblock \doi{10.18653/v1/N19-3005}.

\bibitem[Pennington et~al.(2014)Pennington, Socher, and
  Manning]{embeddings/GloVe}
Jeffrey Pennington, Richard Socher, and Christopher~D. Manning.
\newblock Glove: Global vectors for word representation.
\newblock In \emph{Proceedings of the 2014 Conference on Empirical Methods in
  Natural Language Processing, {EMNLP} 2014, October 25-29, 2014, Doha, Qatar,
  {A} meeting of SIGDAT, a Special Interest Group of the {ACL}}, pp.\
  1532--1543, 2014.
\newblock \doi{10.3115/v1/d14-1162}.
\newblock URL \url{https://doi.org/10.3115/v1/d14-1162}.

\bibitem[Peters et~al.(2018)Peters, Neumann, Iyyer, Gardner, Clark, Lee, and
  Zettlemoyer]{embeddings/Elmo}
Matthew~E. Peters, Mark Neumann, Mohit Iyyer, Matt Gardner, Christopher Clark,
  Kenton Lee, and Luke Zettlemoyer.
\newblock Deep contextualized word representations.
\newblock In \emph{Proceedings of the 2018 Conference of the North American
  Chapter of the Association for Computational Linguistics: Human Language
  Technologies, {NAACL-HLT} 2018, New Orleans, Louisiana, USA, June 1-6, 2018,
  Volume 1 (Long Papers)}, pp.\  2227--2237, 2018.
\newblock \doi{10.18653/v1/n18-1202}.
\newblock URL \url{https://doi.org/10.18653/v1/n18-1202}.

\bibitem[Plank et~al.(2016)Plank, S{\o}gaard, and Goldberg]{noise/plank}
Barbara Plank, Anders S{\o}gaard, and Yoav Goldberg.
\newblock Multilingual part-of-speech tagging with bidirectional long
  short-term memory models and auxiliary loss.
\newblock In \emph{Proceedings of the 54th Annual Meeting of the Association
  for Computational Linguistics (Volume 2: Short Papers)}, pp.\  412--418,
  Berlin, Germany, August 2016. Association for Computational Linguistics.
\newblock \doi{10.18653/v1/P16-2067}.
\newblock URL \url{https://www.aclweb.org/anthology/P16-2067}.

\bibitem[Radford et~al.(2019)Radford, Wu, Child, Luan, Amodei, and
  Sutskever]{embeddings/GPT2}
Alec Radford, Jeff Wu, Rewon Child, David Luan, Dario Amodei, and Ilya
  Sutskever.
\newblock Language models are unsupervised multitask learners.
\newblock 2019.

\bibitem[Rahimi et~al.(2019)Rahimi, Li, and Cohn]{ner/Rahimi2019Massively}
Afshin Rahimi, Yuan Li, and Trevor Cohn.
\newblock Massively multilingual transfer for {NER}.
\newblock In \emph{Proceedings of the 57th Annual Meeting of the Association
  for Computational Linguistics}, pp.\  151--164, Florence, Italy, July 2019.
  Association for Computational Linguistics.
\newblock \doi{10.18653/v1/P19-1015}.
\newblock URL \url{https://www.aclweb.org/anthology/P19-1015}.

\bibitem[Ratinov \& Roth(2009)Ratinov and Roth]{distant/Ratinov2009}
Lev Ratinov and Dan Roth.
\newblock Design challenges and misconceptions in named entity recognition.
\newblock In \emph{Proceedings of the SIGNLL Conference on Computational
  Natural Language Learning (CoNLL 2009)}, 2009.

\bibitem[Ratner et~al.(2019)Ratner, Bach, Ehrenberg, Fries, Wu, and
  R{\'e}]{distant/Ratner2019Snorkel}
Alexander Ratner, Stephen~H. Bach, Henry Ehrenberg, Jason Fries, Sen Wu, and
  Christopher R{\'e}.
\newblock Snorkel: rapid training data creation with weak supervision.
\newblock \emph{The VLDB Journal}, Jul 2019.
\newblock ISSN 0949-877X.
\newblock \doi{10.1007/s00778-019-00552-1}.
\newblock URL \url{https://doi.org/10.1007/s00778-019-00552-1}.

\bibitem[Ratner et~al.(2016)Ratner, De~Sa, Wu, Selsam, and
  R\'{e}]{distant/DataProgramming}
Alexander~J Ratner, Christopher~M De~Sa, Sen Wu, Daniel Selsam, and Christopher
  R\'{e}.
\newblock Data programming: Creating large training sets, quickly.
\newblock In D.~D. Lee, M.~Sugiyama, U.~V. Luxburg, I.~Guyon, and R.~Garnett
  (eds.), \emph{Advances in Neural Information Processing Systems 29}, pp.\
  3567--3575. Curran Associates, Inc., 2016.

\bibitem[Rehbein \& Ruppenhofer(2017)Rehbein and Ruppenhofer]{noise/Rehbein17}
Ines Rehbein and Josef Ruppenhofer.
\newblock Detecting annotation noise in automatically labelled data.
\newblock In \emph{Proceedings of the 55th Annual Meeting of the Association
  for Computational Linguistics (Volume 1: Long Papers)}, pp.\  1160--1170,
  Vancouver, Canada, July 2017. Association for Computational Linguistics.
\newblock \doi{10.18653/v1/P17-1107}.
\newblock URL \url{https://www.aclweb.org/anthology/P17-1107}.

\bibitem[Sanh et~al.(2019)Sanh, Debut, Chaumond, and
  Wolf]{embeddings/DistilBERT}
Victor Sanh, Lysandre Debut, Julien Chaumond, and Thomas Wolf.
\newblock Distilbert, a distilled version of {BERT:} smaller, faster, cheaper
  and lighter.
\newblock \emph{CoRR}, abs/1910.01108, 2019.
\newblock URL \url{http://arxiv.org/abs/1910.01108}.

\bibitem[Str{\"{o}}tgen et~al.(2018)Str{\"{o}}tgen, Gertz, Hirst, and
  Huang]{distant/Stroetgen18Time}
Jannik Str{\"{o}}tgen, Michael Gertz, Graeme Hirst, and Ruihong Huang.
\newblock Domain-sensitive temporal tagging.
\newblock \emph{Computational Linguistics}, 44\penalty0 (2), 2018.

\bibitem[Tjong Kim~Sang \& De~Meulder(2003)Tjong Kim~Sang and
  De~Meulder]{data/CoNLL03}
Erik~F. Tjong Kim~Sang and Fien De~Meulder.
\newblock Introduction to the {C}o{NLL}-2003 shared task: Language-independent
  named entity recognition.
\newblock In \emph{Proceedings of the Seventh Conference on Natural Language
  Learning at {HLT}-{NAACL} 2003}, pp.\  142--147, 2003.

\bibitem[Veit et~al.(2017)Veit, Alldrin, Chechik, Krasin, Gupta, and
  Belongie]{noise/Veit17}
Andreas Veit, Neil Alldrin, Gal Chechik, Ivan Krasin, Abhinav Gupta, and
  Serge~J. Belongie.
\newblock Learning from noisy large-scale datasets with minimal supervision.
\newblock In \emph{Proceedings of the IEEE Conference on Computer Vision and
  Pattern Recognition (CVPR)}, 2017.
\newblock \doi{10.1109/CVPR.2015.7298885}.

\bibitem[Wang et~al.(2019)Wang, Liu, Li, Yang, and Li]{noise/Wang19}
Hao Wang, Bing Liu, Chaozhuo Li, Yan Yang, and Tianrui Li.
\newblock Learning with noisy labels for sentence-level sentiment
  classification.
\newblock In \emph{Proceedings of the 2019 Conference on Empirical Methods in
  Natural Language Processing and the 9th International Joint Conference on
  Natural Language Processing, {EMNLP-IJCNLP} 2019}, pp.\  6285--6291, 2019.
\newblock \doi{10.18653/v1/D19-1655}.

\bibitem[Xiao et~al.(2015)Xiao, Xia, Yang, Huang, and
  Wang]{data/Xiao2015Clothing1M}
Tong Xiao, Tian Xia, Yi~Yang, Chang Huang, and Xiaogang Wang.
\newblock Learning from massive noisy labeled data for image classification.
\newblock In \emph{Proceedings of the IEEE conference on computer vision and
  pattern recognition}, pp.\  2691--2699, 2015.
\newblock \doi{10.1109/CVPR.2015.7298885}.

\bibitem[Yang et~al.(2018)Yang, Chen, Li, He, and Zhang]{Yang18Distantly}
Yaosheng Yang, Wenliang Chen, Zhenghua Li, Zhengqiu He, and Min Zhang.
\newblock Distantly supervised ner with partial annotation learning and
  reinforcement learning.
\newblock In \emph{Proceedings of COLING 2018}, 2018.
\newblock URL \url{https://www.aclweb.org/anthology/C18-1183}.

\bibitem[Zhang et~al.(2018)Zhang, Lin, Pan, Lu, May, Knight, and
  Ji]{ner/Zhang18ELISA}
Boliang Zhang, Ying Lin, Xiaoman Pan, Di~Lu, Jonathan May, Kevin Knight, and
  Heng Ji.
\newblock Elisa-edl: A cross-lingual entity extraction, linking and
  localization system.
\newblock In \emph{Proceedings of NAACL-HLT 2018: Demonstrations}, 2018.
\newblock \doi{10.18653/v1/N18-5009}.
\newblock URL \url{http://aclweb.org/anthology/N18-5009}.

\end{thebibliography}
\bibliographystyle{iclr2020_conference}

\end{document}